\documentclass[runningheads]{llncs}
\usepackage[T1]{fontenc}
\usepackage{hyperref} 
\usepackage{cleveref} 
\usepackage{booktabs} 
\usepackage{xspace} 
\usepackage{subcaption} 
\usepackage{fancyhdr} 
\usepackage{tabulary} 
\usepackage{tcolorbox} 
\usepackage{listings} 

\NewDocumentCommand{\rot}{O{-20} O{1em} m}{\makebox[#2][r]{\hspace{0.5em}\rotatebox[origin=r]{#1}{#3}}}

\newcommand{\LKGBf}{LLM-KG-Bench framework\xspace} 
\newcommand{\LKGBF}{LLM-KG-Bench Framework\xspace} 

\makeatletter 
\newcommand{\printfnsymbol}[1]{%
  \textsuperscript{\@fnsymbol{#1}}%
}
\makeatother 

\definecolor{formalshade}{rgb}{0.95,0.95,1}
\definecolor{darkblue}{rgb}{0.0, 0.0, 0.55}
\newtcolorbox{chatBoxPrompt}{
    colback = formalshade, 
    colframe = darkblue, 
    boxrule = 0pt,
    halign = flush left, 
    fontupper = \scriptsize, 
    size = fbox, 
    leftrule = 4pt, 
}
\newtcolorbox{chatBoxLlm}{
    colback = formalshade, 
    colframe = darkblue, 
    boxrule = 0pt,
    halign = flush left, 
    fontupper = \scriptsize, 
    size = fbox, 
    rightrule = 4pt 
}

\begin{document}
\title{LLM-KG-Bench 3.0: A Compass for Semantic Technology Capabilities in the Ocean of LLMs}

\titlerunning{LLM-KG-Bench Framework 3.0}
%
\author{Lars-Peter Meyer\inst{1,2} \orcidID{0000-0001-5260-5181} \and
Johannes Frey\thanks{equal contribution} \inst{1,3} \orcidID{0000-0003-3127-0815} \and
Desiree Heim\printfnsymbol{1} \inst{4} \orcidID{0000-0003-4486-3046} \and
Felix Brei\printfnsymbol{1} \inst{1} \orcidID{0009-0008-5245-6655} \and
Claus Stadler\inst{1} \orcidID{0000-0001-9948-6458} \and
Kurt Junghanns\inst{1}\orcidID{0000-0003-1337-2770} \and
Michael Martin\inst{1,2} \orcidID{0000-0003-0762-8688}}
\authorrunning{L.-P. Meyer et al.}
%
\institute{InfAI, Leipzig, Germany, \email{lpmeyer@infai.org}, \url{https://infai.org}\and
Chemnitz University of Technology, Germany, \url{https://tu-chemnitz.de/} \and
Leipzig University, Germany, \url{https://uni-leipzig.de} \and
DFKI, Kaiserslautern, Germany, \url{https://www.dfki.de}}

\maketitle              

\begin{picture}(0,0)
  \put(\dimexpr\paperwidth-3in\relax, -1.2in){%
    \makebox(0,0)[r]{\rotatebox{90}{\LARGE peer reviewed and to appear in proceedings of \href{https://2025.eswc-conferences.org/conference-proceedings/}{ESWC 2025 Resources Track}}}%
  }
\end{picture}

\begin{abstract}
Current Large Language Models (LLMs) can assist developing program code beside many other things, but can they support working with Knowledge Graphs (KGs) as well?
Which LLM is offering the best capabilities in the field of Semantic Web and Knowledge Graph Engineering (KGE)?
Is this possible to determine without checking many answers manually?
The LLM-KG-Bench framework in Version 3.0 is designed to answer these questions.
It consists of an extensible set of tasks for automated evaluation of LLM answers and covers different aspects of working with semantic technologies.

In this paper the LLM-KG-Bench framework is presented in Version 3 along with a dataset of prompts, answers and evaluations generated with it and several state-of-the-art LLMs.
Significant enhancements have been made to the framework since its initial release, including an updated task API that offers greater flexibility in handling evaluation tasks, revised tasks, and extended support for various open models through the \textit{vllm} library, among other improvements. 
A comprehensive dataset has been generated using more than 30 contemporary open and proprietary LLMs, enabling the creation of exemplary model cards that demonstrate the models' capabilities in working with RDF and SPARQL, as well as comparing their performance on Turtle and JSON-LD RDF serialization tasks.

\begin{description}
    \item[Resource type:] \textbf{evaluation benchmark framework}
    \item[License:] MPL 2.0
    \item[DOI:] \href{https://doi.org/10.5281/zenodo.15100803}{DOI:10.5281/zenodo.15100803}
    \item[URL:] \url{https://github.com/AKSW/LLM-KG-Bench}
\end{description}

\keywords{LLM \and RDF \and SPARQL \and Knowledge Graph \and LLM Evaluation}
\end{abstract}
\section{Introduction}
In the rapidly developing field of Large Language Models (LLMs), it is difficult to keep up with the latest developments and put them into context of prior work.
Several LLMs are released every month and some of them are advertised as \emph{"better"}, \emph{"faster"}, \emph{"cheaper"} or with \emph{better "reasoning capabilities"}. 
With the work on our benchmarking framework \emph{LLM-KG-Bench}, we are particularly interested in making it possible to assess and compare LLMs by their capabilities to cope with Semantic Web technology. 
The framework features support for open source and top-ranking commercial LLMs and includes a set of highly relevant tasks specific to Knowledge Graph Engineering (KGE). 
For instance, there are specialized tasks related to SPARQL and RDF serialization.
In this work, we present Version 3.0 of LLM-KG-Bench. 
Our latest advancements offer the following contributions:
\begin{itemize}
\item A major update of the task api which makes writing of new task easier as the overhead is reduced and the framework can handle the evaluation orchestration.
\item An RDF repair task where the goal is to detect and fix errors across several RDF serialization formats, such as Turtle, JSON-LD, N-Triples.
\item Improved analytics and visualization: Combined scores can be computed and visualized in a capability compass for task categories such as \emph{RDF syntax}, \emph{RDF analytics}, \emph{SPARQL syntax}, \emph{SPARQL semantics} or \emph{brevity}.
\item Support for encrypted task data to prevent test data leakage into LLM training data.
\item A new connector for \emph{vllm}\footnote{\url{https://github.com/vllm-project/vllm}}, a popular high throughput LLM serving framework.
\end{itemize}

The extensions and refinement of the framework were guided by making the KGE related comparison of LLMs easier and broader in aspects of LLM size and task areas covered.

The remainder of this work is structured as follows: In~\Cref{sec:related-work} we present related work. An overview of our LLM-KG-Bench system and its latest improvements is given in \Cref{sec:resource-description}. In~\Cref{sec:inUse} we apply the framework to create a big dataset of evaluation results for more than 30 open and proprietary LLMs.
\Cref{sec:conclusions} concludes this work and points out directions for future work.

\section{Related Work}
\label{sec:related-work}

In order to explore and navigate the vast amount of LLMs, there are several LLM leaderboards, which rank LLMs based on a selection of benchmarks or workloads.
For commercial (and a set of open) models, the \textbf{Chatbot Arena}\footnote{Leaderboard: \url{https://huggingface.co/spaces/lmarena-ai/chatbot-arena-leaderboard}} \cite{chiang2024ChatbotArena} is popular.
While it lists scores for \textbf{MMLU}\footnote{Description: \url{https://crfm.stanford.edu/2024/05/01/helm-mmlu.html}} and \textbf{MT-bench}\cite{mt-bench}, it also calculates its own arena-score which is based on arbitrary tasks, that are processed by two models side-by-side and then evaluated by the same user voting for their preferred answer.
For open models the \textbf{OpenLLM-Leaderboard} \cite{open-llm-leaderboard-v2} provides the most exhaustive list of benchmark results with over 2,000 tested models, and with scores for IfEval, BBH, MATH, GPQU, MUSR, MMLU, and a carbon dioxide emission estimate.
HELM\footnote{Leaderboard: \url{https://crfm.stanford.edu/helm/lite/latest/\#/leaderboard}} \cite{Liang2023HolisticEvaluationLanguageModels} comprises the most exhaustive list, including also domain-specific tests like LegalBench and MedQA.
In contrast to a set of other leaderboards that just collect published or reported test results, those leaderboards provide evaluation as a service.
While a comparison of the individual benchmark suites is out of scope of this paper, we see in all of them a gap in addressing Knowledge Graph engineering (KGE) tasks.
We also see a gap in a benchmark execution framework that helps to deal with the particularities of RDF and KG-related workloads (format parsing, syntax check feed back loops, execution and evaluation of queries towards KGs, etc.)
While Big Bench\cite{srivastava2023imitationGameBigBench} was an initial inspiration for the LLM-KG-Bench framework, and we tried to be compatible in the beginning, we realized that the Task API was not sufficient for our KGE benchmarking efforts.
Both HELM and BigBench have a strong focus on multiple choice tasks and use scores based on string or document similarities.
In contrast, the LLM-KG-Bench framework focuses on the syntactically and semantically correct generation of RDF (i.e. in Turtle) and SPARQL.
The LLM-KG-Bench framework aims to reduce complexity and technological burdens to create, execute, evaluate, and analyze KG-related tasks.

In the area of benchmarking coding capabilities, we observe, at a conceptual level, characteristics (e.g., with respect to output format requirements, instruction complexity, and response evaluation strategies) that are more closely related to KGE capabilities benchmarking. 
In this domain, several leaderboards exist.
The \textbf{Big Code Models Leaderboard}\footnote{\url{https://huggingface.co/spaces/bigcode/bigcode-models-leaderboard}} evaluates over 60 base models using the HumanEval and MultiPL-E configured for Java, Javascript, and CPP. 
The \textbf{EvalPlus Leaderboard}\footnote{\url{https://evalplus.github.io/leaderboard.html}} ranks more than 100 models using HumanEval and the Mostly Basic Python Programming (MBPP) Benchmark. 
These datasets combine human-written programming problems with basic Python challenges to assess coding proficiency. 
The \textbf{CanAiCode Leaderboard}\footnote{\url{https://huggingface.co/spaces/mike-ravkine/can-ai-code-results}} focuses on pro\-gram\-ming-re\-la\-ted tasks, benchmarking more than 300 models using the custom CanAICode Benchmark. 
This test suite is specifically designed for testing small text-to-code LLMs with less complex tasks compared to HumanEval and MBPP.
However, these efforts are very specialized towards coding and as such also hard to adopt for our use case.

In literature, many efforts\footnote{Updated overview: \url{https://github.com/zjukg/KG-LLM-Papers}} explore the combination of LLMs and KGs\cite{PanUnifyingLLMsAndKGs}. 
Several of them are evaluating the application of LLMs for KG related tasks.
However, these LLM evaluations are often focused on a very specific problem in a specific task area like Text to RDF (e.g. \cite{Mihindukulasooriya2023Text2KGBench,Zhu2023LlmsKgConstructionReasoning})
or Knowledge Graph Question Answering (KGQA, e.g. \cite{Usbeck2019Benchmarkingquestionanswering})
or Text to SPARQL (e.g. \cite{Kovriguina2023SPARQLGENOP,Zahera2024GeneratingSPARQLNatural})
or RML generation (e.g. \cite{hofer2022towards}).
Unfortunately, many of the evaluations in these articles were conducted manually.
This comes with the problem of not being able to scale those evaluations to more repetitions and more or newer models.
In case an automated evaluation has been performed, the underlying code usually lacks adaptability to encompass new models or task variations to be executed and analyzed.
A benchmarking effort, that is related to our interest in studying the JSON-LD capabilities, is \textbf{StructuredRAG} \cite{shorten2024structuredragjsonresponseformatting}. 
It consists of six tasks designed to assess LLM capabilities in following response format instructions according to JSON templates.
\Cref{tab:otherLlmEvaluationApproaches} compares several LLM evaluation approaches.

\begin{table}[tb]
     \caption{
        Comparison of some of the LLM evaluation approaches mentioned here.
        Best values are marked with bold font.
        Only the \LKGBf combines automatic evaluation with several Knowledge Graph Engineering (KGE) topics and many LLMs covered.
    }
    \centering \scriptsize
    \begin{tabular}{lccc}
        \toprule
         &  LLMs covered&  KGE topics& eval. type\\
        \cmidrule{2-4}
         BIG Bench\cite{srivastava2023imitationGameBigBench}& \textbf{many}&  no& automatic\\
         HELM\cite{Liang2023HolisticEvaluationLanguageModels}& \textbf{many}&  no& automatic\\
         Chatbot Arena\cite{chiang2024ChatbotArena}& \textbf{many}&  no& crowd\\
         ChatGPT KG Experiments\cite{Meyer2023ExperimentsWithChatGPT} & GPT3.5, GPT4 & \textbf{several} & manual\\
         Text2KgBench\cite{Mihindukulasooriya2023Text2KGBench} &  Vicuna, Alpaca-LoRA & Text2KG & automatic\\
         AutoKG\cite{Zhu2023LlmsKgConstructionReasoning} & 4 & Text2KG, Reasoning & manual\\
         SparqlGen\cite{Kovriguina2023SPARQLGENOP} & GPT3 & Text2Sparql & manual\\
         \textbf{LLM-KG-Bench 3} & \textbf{many} & \textbf{several} & automatic\\
        \bottomrule
    \end{tabular}
   
    \label{tab:otherLlmEvaluationApproaches}
\end{table}

The LLM-KG-Bench framework has been described and applied in several publications. 
The following section provides a brief overview.
The LLM-KG-Bench framework was initially introduced in \cite{Meyer2023DevelopingScalableBenchmark}, featuring three basic tasks (Version 1.0).
Version 1.1 expanded the framework to include five tasks, incorporating evaluations of the Turtle capabilities of both open and proprietary LLMs, as detailed in \cite{Frey2023Turtle}. 
In 2023, Version 1.2 collected results for various proprietary LLM versions, as described in \cite{Frey2024AssessingEvolutionLLM}. 
The framework was further enhanced to support task-based dialogues with LLMs and introduced a re-evaluation mode, enabling task evaluations to be rerun using previously generated task data and responses to the same prompts.
In Version 2.0, numerous additional SPARQL tasks and task parameterizations were integrated. 
This version was also utilized to evaluate the SPARQL capabilities of various proprietary LLMs, as detailed in \cite{Meyer2024AssessingSparqlCapabilititesLLM}. 
Building upon this basis, Version 3 of the LLM-KG-Bench framework introduces (a) an expanded task list, (b) supports encrypted task data, (c) includes a major update to the task API, and (d) extends compatibility to more models through additional model connectors.

\section{Resource Description}
\label{sec:resource-description}

Benchmarking LLMs involves significant time and financial costs plus organizational effort, and the evaluation process can often be imprecise.
LLM-KG-Bench is designed to simplify the creation of KG-related assessments while providing a foundational infrastructure for further development.
Its main features are:
\begin{itemize}
    \item \textbf{Modular and Extensible Framework:} Supports automated evaluation tasks using a comprehensive set of KG-extraction and evaluation-related helper methods.
    \item \textbf{Built-in Correction Cycles:} Implements dialogue-based correction cycles, enabling LLMs to revise previous mistakes.
    \item \textbf{Data Security:} Supports encryption of task data to prevent test data leakage into LLM training datasets.
    \item \textbf{Task Management:} Manages task configurations, evaluation orchestration, logging, and result persistence.
    \item \textbf{Result Analysis and Visualization:} Provides built-in tools for analyzing and visualizing evaluation results.
    \item \textbf{Broad Model Support:} Includes connectors for many contemporary LLMs.
    \item \textbf{Open-Source Codebase:} The framework is published as open source and welcomes extensions and community contributions.
\end{itemize}

The main architecture is described in Figure 1 of Meyer et al.\cite{Meyer2023DevelopingScalableBenchmark}.
In the following sections, we describe the basic concepts and infrastructure of LLM-KG-Bench framework in greater detail.

\subsection{Main Concepts of the \LKGBF}

The \LKGBf is build around some main concepts we want to describe here.

\paragraph{Evaluation Tasks:}
The \emph{evaluation tasks} are the main building block of a \emph{benchmark} and automatically evaluate the LLM answers.
For the \emph{prompt-answer-evaluate loop} the tasks provide the prompt and evaluation functionality.

\paragraph{Task Classes, Parametrized Tasks and Task Case Entries:}
Tasks are organized in \emph{task classes}.
Some task classes can be parametrized with task class specific \emph{task parameters} resulting in \emph{parametrized task classes}.

\begin{figure}[tb]
    \centering
    
    \begin{subfigure}[t]{0.4\textwidth}
        \includegraphics[width=\linewidth]{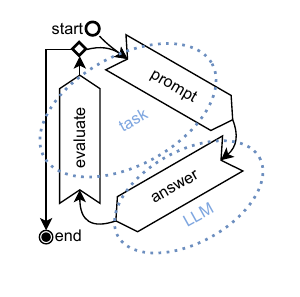}
        \caption{
            The Prompt-Answer-Evaluate loop for the task - LLM interaction as organized by the framework.
            Prompting and evaluation is covered by the task, the answer is generated by the LLM.
        }
        \label{fig:PromptAnswerEvalLoop}
    \end{subfigure}
    \hfill
    \begin{subfigure}[t]{0.58\textwidth}
        \includegraphics[width=\linewidth]{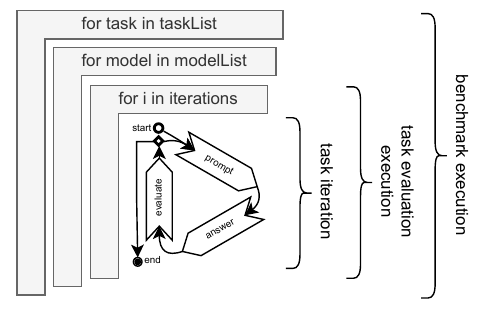}
        \caption{Different execution scopes: task iterations (includes one to many cycles of the prompt-answer-evaluate loop), task execution (includes all iterations), benchmark execution (includes all task executions for all combinations of tasks and LLMs defined for all iterations).}
        \label{fig:TaskIterBenchExec}
    \end{subfigure}
    \caption{Overview of the evaluation workflow and execution scopes.
    }
\end{figure}

\paragraph{Prompt-Answer-Evaluate Loop:}
The evaluation of LLMs is based on dialogues, consisting of prompts and answers.
The prompt-answer-evaluate loop starts with the generation of an initial prompt that is sent to an LLM. 
In the next step, the produced answer is evaluated.
Based on the evaluation result, the framework can decide to start a new prompt-answer-evaluate  round or stop the dialogue.
The idea is to make use of the chat capability of modern LLMs and their bigger supported context size in order to get the answer closer to the correct one.
The structure of these loops is shown in \cref{fig:PromptAnswerEvalLoop} and an example dialogue is given in \cref{fig:DialogExample}.

\paragraph{LLM Connectors:}
The \emph{LLM connectors} offer a consistent abstraction layer to interact with various supported LLMs.
Several \emph{LLM connector classes} are implemented in the \LKGBf as described in \cref{sec:ModelConnectors}.
They can get parametrized to abstract specific LLMs.

\paragraph{Task Evaluation Iterations:}
We name one task evaluation loop consisting of one or more prompt-answer-evaluate rounds a \emph{task evaluation iteration}, see also \cref{fig:TaskIterBenchExec}.

\paragraph{Task Executions:}
Since LLM answers are generated probabilistically, a configurable number of task iterations is executed, collectively forming a \emph{task evaluation execution} for a specific task and a particular LLM, see also \cref{fig:TaskIterBenchExec}.

\paragraph{Benchmark Executions:}
A \emph{benchmark execution} consists of all task executions for all combinations of tasks and models defined in the configuration, see also \cref{fig:TaskIterBenchExec}.

\paragraph{List Tasks and Task Case Entries:}
\emph{List tasks} have a list of \emph{task case entries}, where each entry defines a distinct exercise resulting in a specific prompt and expected answer.
For each \emph{task iteration} one \emph{task case entry} is selected from this list.
All \emph{task case entries} are evaluated by the same \emph{list task}.

\paragraph{Benchmark Configuration:}
A \emph{benchmark configuration} specifies the tasks and models to be included in a benchmark run together with the number of iterations per task execution.

\paragraph{Execution Configuration}
A \emph{benchmark configuration} can be executed as a whole or with a selection of tasks and models defined by command line parameters.

\paragraph{Result Reevaluation}
With result reevaluation existing task model interaction data can be feed into the tasks evaluation code.
This could help for example to get updated results with updated evaluation code without new LLM interactions.

\subsection{Task API}
\begin{figure}[tb]
    \centering
    \includegraphics[width=\linewidth]{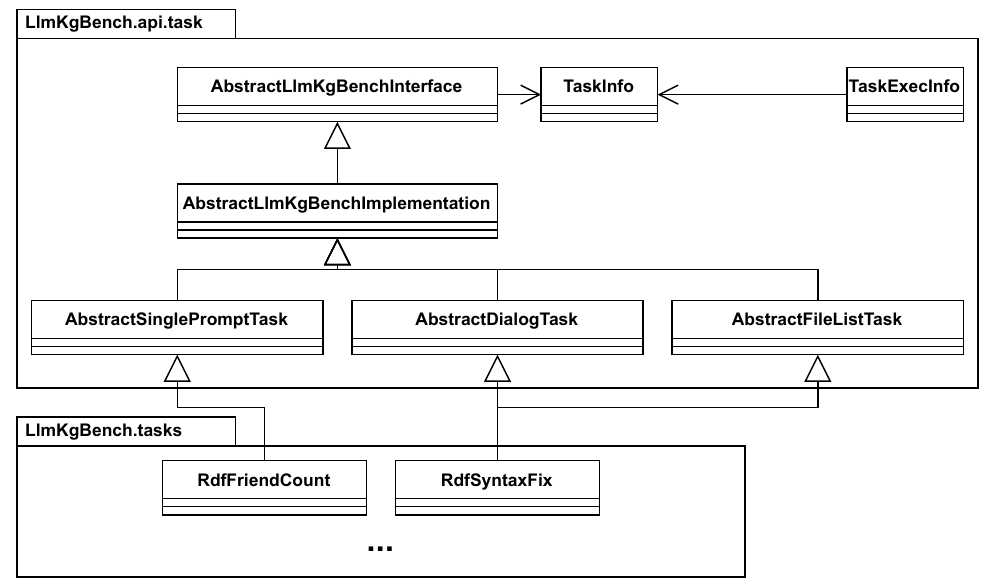}
    \caption{
        UML class diagram of the Task API and its reference by some example tasks.
        All task classes implement the \emph{AbstractLlmKgBenchInterface} via an inheritance connection.
        A task can be described with a \emph{TaskInfo} object.
        A \emph{TaskExecutionInfo} references this \emph{TaskInfo} object for the documentation.
    }
    \label{fig:UML-ClassDiagram-Tasks}
\end{figure}

Tasks are implemented following the task API as a common interface between tasks, framework and model connectors.
In \LKGBf Version~3, a major update of the Task API was introduced together with new helper classes.
Figure \ref{fig:UML-ClassDiagram-Tasks} shows the UML class diagram of the new Task API.

Benchmark tasks in LLM-KG-Bench do implement the interface \emph{AbstractLlmKgBenchTaskInterface} which enables a rough compatibility with the BigBench task classes.
As supplement, it defines methods for the Evaluate and Prompt part of the Prompt-Answer-Evaluate loop as well as methods for the serialization and deserialization of tasks.

Here, the following methods are especially important:
\begin{description}
    \item[getNextPrompt:] combines an evaluation and prompting step.
        If no new prompt is generated the prompt-answer-evaluate loop ends.
    \item[finalizeEvaluation:] is called at the end of the prompt-answer-evaluate loop and creates a final evaluation result for this task evaluation iteration.
    \item[condenseTaskData:] creates a serializable representation of this concrete task case entry.
        This offers the possibility for later continuation or reevaluation.
    \item[createTaskFromCondensedData:] initializes a task from the representation given by \emph{condenseTaskData}
\end{description}

The abstract implementation \emph{AbstractLlmKgBenchTaskClass} helps to reduce redundant code and eases the concrete task implementation.
For the two main variations of tasks, single-prompt tasks and dialogue-tasks, specialized abstract classes are provided.
Tasks which store their task data in an encrypted file can benefit from the abstract class AbstractFileListTaskImplementation.

This new task API offers more granularity for the interaction of the \LKGBf with the tasks compared to the interface used in prior versions of the \LKGBf and BigBench.
The central framework logic orchestrates the prompt-answer-evaluate loop(\cref{fig:PromptAnswerEvalLoop}).
The new task API gives the central framework logic more flexibility in the orchestration, reduces error possibilities and reduces repeated code in tasks.

\subsection{Model Connectors and Supported Models}
\label{sec:ModelConnectors}
Model connectors are responsible for offering standardized APIs to LLMs.
They are defined similar to the BIG bench model class.
The main method offered is \verb|generate_text(inputs: Union[str,List[str]], ...)->str|, taking a single prompt or dialogue and returning the LLMs answer. 

The \LKGBf offers several model connectors:
\begin{description}
    \item[OpenAI / ChatGPT:] Connector for OpenAI-compatible LLMs like GPT-3.5, GPT-4, GPT-4t, GPT-4o and GPT-o1 via the OpenAI python library\footnote{Repository: \url{https://github.com/openai/openai-python}} and REST API\footnote{API description: \url{https://platform.openai.com/docs/api-reference/chat}}.
    Many other LLM providers offer a compatible REST API as well.
    They can be integrated with the new endpoint parameter.
    \item[Google / Gemini:] Connector for LLMs from Google like Gemini 1.5 or Gemini 2.0 via the Google python library\footnote{Repository: \url{https://github.com/google-gemini/generative-ai-python}} and REST API\footnote{API description: \url{https://ai.google.dev/api}}.
    \item[Anthropic / Claude:] Connector for LLMs from Anthropic from Claude 1.0 to Claude 3.5.
    The connector uses the Anthropic REST API\footnote{API description: \url{https://docs.anthropic.com/en/api}} using the offered python library\footnote{Repository: \url{https://github.com/anthropics/anthropic-sdk-python}}.
    \item[vLLM]: Runtime for self-hosted LLMs \footnote{webpage: \url{https://docs.vllm.ai}}\cite{kwon2023VllmEfficientMemoryManagement}.
    This library is compatible to many open LLMs and enables serving and inferencing them. 
\end{description}

\subsection{Benchmark Tasks}
\label{sec:LlmKgBenchTaskList}

Several tasks are implemented in \LKGBf as described in several articles \cite{Meyer2023DevelopingScalableBenchmark,Frey2023Turtle,Frey2024AssessingEvolutionLLM,Meyer2024AssessingSparqlCapabilititesLLM}. This includes especially the following task classes:
\begin{itemize}
    \itemindent=-1.7em 
    \item[] RDF related: 
        \begin{description}
            \item[FactExtractStatic:] asks the LLM to extract facts from a given textual fact sheet and create a corresponding Turtle KG
            \item[RdfConnectionExplainStatic (extended):] asks the LLM to find a connection between two nodes in a small RDF graph
            \item[RdfFriendCount (extended):] for a simple dynamically computed RDF graph the LLM should find the node with the most incoming edges
            \item[RdfSyntaxFixList (new):] present the LLM a RDF document with syntax errors and asks for a fixed document
            \item[TurtleSampleGeneration:] asks the LLM to create a small graph of foaf:Person objects connected with foaf:knows edges
        \end{description}
    \item[] SPARQL SELECT query related:
        \begin{description}
            \item[Sparql2AnswerList:] presents the LLM a small KG and SPARQL SELECT query and asks for the expected result set of the query
            \item[Text2AnswerList:] presents the LLM a small KG and textual question and asks for the expected result set of the question
            \item[Text2SparqlList:] asks the LLM to translate a given textual question into a SPARQL SELECT query for a given KG
            \item[SparqlSyntaxFixingList:] presents a SPARQL SELECT query with syntax errors and asks fo a fixed query
        \end{description}
\end{itemize}
Many of this task classes support parameters and there are several variations of the Sparql2AnswerList task class for different benchmark datasets.
Some tasks, especially the SPARQL related ones, are based on existing task data \cite{dubey2017lcQuad2,Meyer2023ExperimentsWithChatGPT,Kovriguina2023SPARQLGENOP,Brei2024LeveragingSmallLanguage,Brei2024Queryfy}.

The prompts are designed in a way that keeps ambiguity as low as possible.
All requirements that we expect the LLM to honor are stated explicitly, e.g. \verb|stick with the original formatting|, or \verb|answer with just one markdown| \verb|fenced code block ... no other text|.
At the same time we avoid LLM specific prompt optimization for a fair comparison across different and future models.

\subsubsection{New RdfSyntaxFixList Task}
The task presents an RDF document with one or two syntax errors together with the related parsing error and asks the LLM to fix the document with few as possible changes, similar to the TurtleErrorsStatic task introduced in the first version of \LKGBf\cite{Meyer2023DevelopingScalableBenchmark}.
But where the TurtleErrorsStatic task is limited to one document and tries to estimate the syntactical correctness of a Turtle document that still contains errors, we decided to take a more generic approach.
When the document returned by the LLM contains still some errors or differs from the expected formatting, the task provides feedback and asks the LLM in another prompt-answer-eval loop to fix it again.
This is repeated for up to three rounds and the documents generated are measured according to several aspects resulting in the following scores:
\begin{description}
    \item[parsableSyntax] $\in [0,1]$: is the document syntax correct?
    \item[contentF1] $\in [0..1]$: f1 measure for the document generated in comparison with the expected content on a normalized triple level.
    This score is computed only if the document syntax is correct.
    \item[strSimilarity] $\in [0..1]$: string similarity of the document generated to the expected result.
    \item[brevity] $\in [0..1]$: is only the document provided or additional surrounding text generated we did not ask for?
    \item[combined] $\in [0..1]$: combined score computed by \\
    $0.1 strSimilarity + 0.2 parsableSyntax + 0.7 contentF1$
    
\end{description}

This is implemented for Turtle, JSON-LD and N-Triples.
The selection of the RDF serialization format is configured with a task parameter \verb|graphFormat|.
The task is based on an encrypted task data file containing five task case entries for each serialization format supported.

Each task case entry is based on variations of the organizational graph with between one and four syntax errors.
The errors could be a missing or additional formatting character, an invalid combination of literal language and type, a misspelled  prefix or a wrong character escaping in a string literal.
An example dialogue for a missing formatting character in Turtle is shown in \cref{fig:DialogExample}.

\begin{figure}
    \begin{chatBoxPrompt}
        \textbf{Bench:} Please fix all syntax errors of the following RDF in turtle syntax.
        Try to stick with the original formatting of the RDF given and only change as few characters as necessary.
        To support automated parsing, please answer with just one markdown fenced code block (start and end with \verb|```|) containing the rdf, no other text.
        \vspace{12pt}
\begin{lstlisting}[aboveskip=0pt,belowskip=0pt,escapechar={\~}]
```turtle
~\dots~
:anne a foaf:Person ; foaf:firstName "Anne"
~\dots~
```
\end{lstlisting} 
        Parsing error message: \verb|at line 7 of <>: Bad syntax|
        \verb| (expected '.' or '}' or ']'| \verb| at end of statement)| \verb| at ^ in:| \dots
    \end{chatBoxPrompt}
    \begin{chatBoxLlm}
        \textbf{LLM:} A dot (.) is missing
    \end{chatBoxLlm}
    \begin{chatBoxPrompt}
        \textbf{Bench:}
        Please correct your answer following the expected structure(exactly one fenced code block with the RDF, no other text).
    \end{chatBoxPrompt}
    \begin{chatBoxLlm}
        \textbf{LLM:}
\begin{lstlisting}[aboveskip=0pt,belowskip=-6pt,escapechar={\~}]
```turtle
~\dots~
:anne a foaf:Person ; foaf:firstName "Anne" .
~\dots~
```
\end{lstlisting}
    \end{chatBoxLlm}
    \caption{
        Example dialogue for the RdfSyntaxFixList task with a missing dot in Turtle syntax.
        Some text left out is marked with "\dots" .
        The LLM's first answer is missing the expected code block with the fixed Turtle which is corrected in the second answer.
    }
    \label{fig:DialogExample}
\end{figure}

\subsubsection{Extended RdfConnectionExplainStatic and RdfFriendCount Tasks}

The task classes RdfConnectionExplainStatic and RdfFriendCount, first introduced in the Version 1.1 of the \LKGBf \cite{Frey2023Turtle}, got extended to support more graph serialization formats.
Instead of just presenting a Turtle graph, they support the four RDF formats Turtle, JSON-LD, RDF/XML and N-Triples now.
The graph format can be configured with a task parameter \verb|graphFormat|.

\subsection{Summary of Updates in Version 3}
In comparison to the earlier versions of the \LKGBf the Version 3 offers especially the following new features:
\begin{itemize}
    \item major update in task API for reduced task code and clean dialogue orchestration
    \item framework support for task data encryption
    \item added vllm connector
    \item new result analysis capability with score aggregation
    \item new plot type spider plot as shown in the capability compass in \autoref{sec:eval}
    \item new task and new task variations with task parameters
    \item some reorganization like modularized model definition
\end{itemize}

\section{Benchmark in Use}
\label{sec:inUse}

To showcase the benchmark, we conducted a broad evaluation of some of the state-of-the-art proprietary and open LLMs with the help of \LKGBf.
All generated data is published in a GitHub Repository and can be accessed with the link given in the "Online Resources" Section at the end of this article to enable further analysis and comparison.

\subsection{Dataset Generation}
We adopted the default configuration to define the tasks and models included in this evaluation.
As a trade-off between resource usage and confidence, we have decided to conduct 20 iterations for the proprietary LLMs and 50 iterations for the open LLMs.

\subsubsection{Selected Tasks}
 The benchmark was executed on the following tasks that are described in \cref{sec:LlmKgBenchTaskList} and in the code repository:
\begin{itemize}
    \item RdfSyntaxFixList: For Turtle, JSON-LD and N-Triples as graph format
    \item RdfConnectionExplainStatic: For Turtle, JSON-LD, RDF/XML and N-Triples as graph format
    \item RdfFriendCount: For Turtle, JSON-LD, RDF/XML and N-Triples as graph format
    \item SparqlSyntaxFixingList
    \item Sparql2AnswerList: For Organizational graph
    \item Text2SparqlList: For Organizational graph, Coypu-Mini and Beastiary
\end{itemize}

\begin{table}[tb!]
    \caption{
        Details for the models selected for the experiment presented here.
        The parameter count of proprietary models is not documented and marked with a question mark (?) here.
    }
    \label{tab:ModelDetails}
    \centering \scriptsize
    \begin{tabular}{lcccc}
    \toprule
    Connector & Family           & Model                       & Parameter            & Context \\
    \midrule
OpenAI    & OpenAI-GPT       & ChatGPT 3.5 turbo           & ?                    & 16k     \\
          &                  & ChatGPT 4o                  & ?                    & 128k    \\
          &                  & ChatGPT 4o-mini             & ?                    & 128k    \\
          &                  & ChatGPT o1                  & ?                    & 128k    \\
          &                  & ChatGPT o1-mini             & ?                    & 128k    \\
    \cmidrule{1-3}
Google    & Google-Gemini    & Gemini 2.0 Flash            & ?                    & 128k-1M \\
          &                  & Gemini 1.5 Pro              & ?                    & 128k-2M \\
          &                  & Gemini 1.5 Flash            & ?                    & 128k-1M \\
    \cmidrule{1-3}
Claude    & Anthropic-Claude & Claude 3.5 Sonnet           & ?                    & 200k    \\
          &                  & Claude 3.5 Haiku            & ?                    & 200k    \\
    \cmidrule{1-3}
vLLM      & Qwen\cite{hui2024qwen25codertechnicalreport,yang2024qwen2technicalreport}             & Qwen2-0.5B-Instruct         & 0.5B                 & 32k     \\
          &                  & Qwen2-1.5B-Instruct         & 1.5B                 & 32k     \\
          &                  & Qwen2-7B-Instruct           & 7B                   & 128k    \\
          &                  & Qwen2-57B-A14B-Instruct     & 57B (active: 14B)    & 64k     \\
          &                  & Qwen2-72B-Instruct          & 72B                  & 128k    \\
          &                  & Qwen2.5-0.5B-Instruct       & 0.5B                 & 32k     \\
          &                  & Qwen2.5-1.5B-Instruct       & 1.5B                 & 32k     \\
          &                  & Qwen2.5-3B-Instruct           & 3B                   & 32k     \\
          &                  & Qwen2.5-7B-Instruct           & 7B                   & 128k    \\
          &                  & Qwen2.5-14B-Instruct          & 14B                  & 128k    \\
          &                  & Qwen2.5-32B-Instruct          & 32B                  & 128k    \\
          &                  & Qwen2.5-72B-Instruct          & 72B                  & 128k    \\
          &                  & Qwen2.5-Coder-32B-Instruct  & 32B                  & 128k    \\
    \cmidrule{2-3}
          & Meta-Llama \cite{grattafiori2024llama3herdmodels}       & Meta-Llama-3-8B-Instruct    & 8B                   & 8k      \\
          &                  & Meta-Llama-3-70B-Instruct   & 70B                  & 8k      \\
          &                  & Llama-3.1-8B-Instruct       & 8B                   & 128K    \\
          &                  & Llama-3.1-70B-Instruct      & 70B                  & 128K    \\
          &                  & Llama-3.2-1B-Instruct       & 1B                   & 128K    \\
          &                  & Llama-3.2-3B-Instruct       & 3B                   & 128K    \\
          &                  & Llama-3.3-70B-Instruct      & 70B                  & 128K    \\
    \cmidrule{2-3}
          & Microsoft-Phi \cite{abdin2024phi3technicalreporthighly}    & Phi-3-mini-128k-instruct    & 3.8B                 & 128k    \\
          &                  & Phi-3-small-128k-instruct   & 7B                   & 128k    \\
          &                  & Phi-3-medium-128k-instruct  & 14B                  & 128k    \\
          &                  & Phi-3.5-mini-instruct       & 3.8B                 & 128k    \\
          &                  & Phi-3.5-MoE-instruct        & 42B (active: 6.6B) & 128k    \\
    \cmidrule{2-3}
          & Infly-OpenCoder \cite{huang2024opencoderopencookbooktoptier}  & OpenCoder-8B-Instruct       & 8B                   & 8k      \\
    \cmidrule{2-3}
          & Deepseek-ai \cite{guo2024deepseekcoderlargelanguagemodel}      & deepseek-coder-33b-instruct & 33B                  & 16k     \\
            \bottomrule
    \end{tabular}
\end{table}

\subsubsection{Selection of Proprietary LLMs}
To get an overview on the current state-of-the-art proprietary models we selected three long-term high-ranked model families from the Chatbot Arena Leaderboard: OpenAI GPT, Google Gemini and Anthropic Claude.
From these families, we selected the current models in various sizes and also included the latest GPT 3.5 for comparability with other results.
The selected models together with their context size are shown in \cref{tab:ModelDetails}

\begin{figure}[tb]
    \centering
    \begin{subfigure}{0.23\textwidth}
        \includegraphics[width=\linewidth]{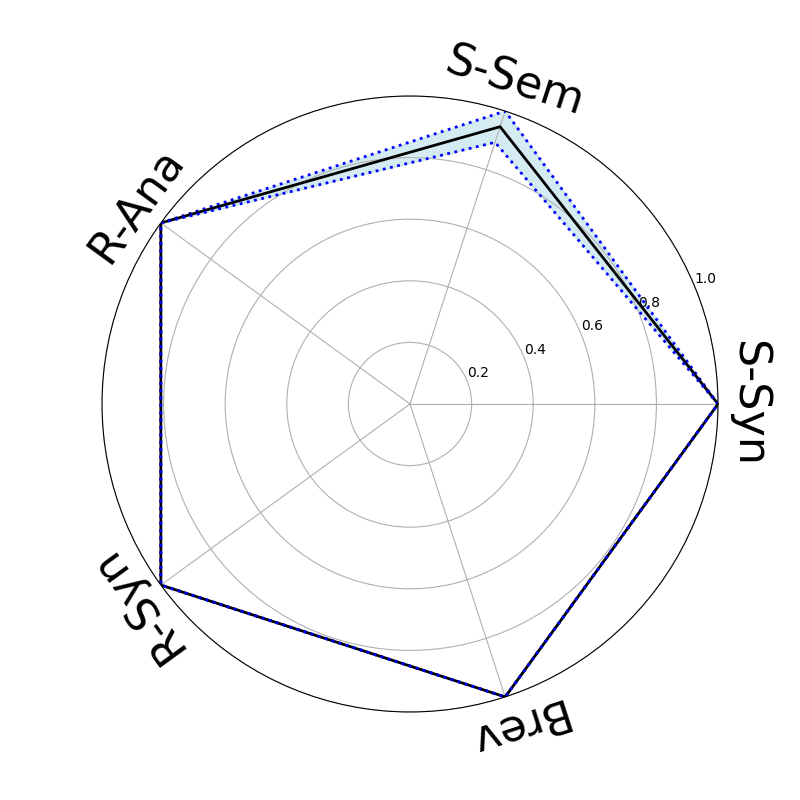}
        \caption{Gemini 2.0 flash}
        \label{fig:compass-Gemini-2-flash}
    \end{subfigure}
    \begin{subfigure}{0.23\textwidth}
        \includegraphics[width=\linewidth]{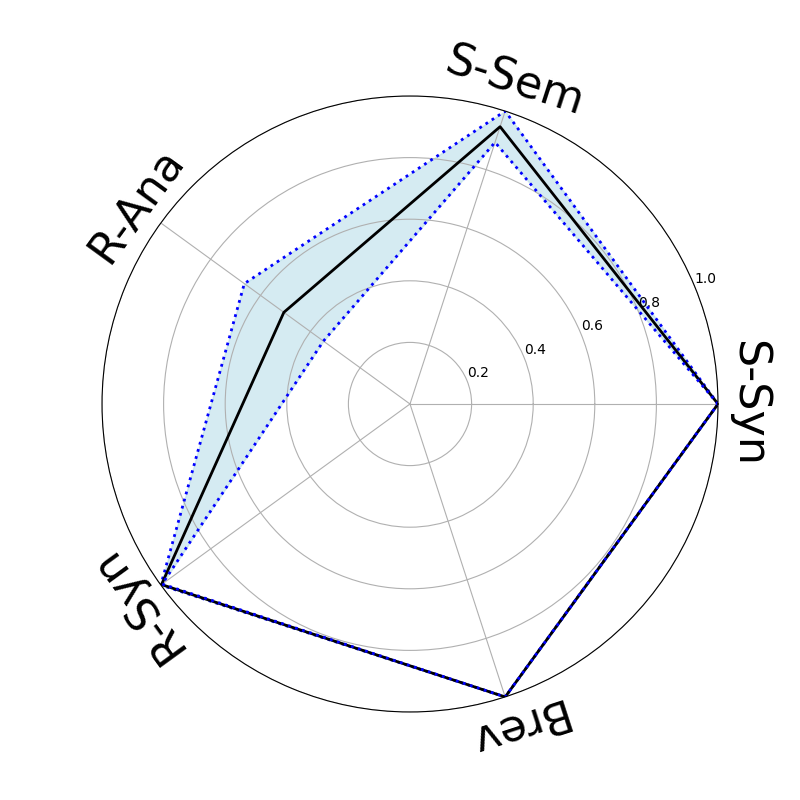}
        \caption{GPT3.5t 24-01}
        \label{fig:compass-GPT-3-5t}
    \end{subfigure}
    \begin{subfigure}{0.23\textwidth}
        \includegraphics[width=\linewidth]{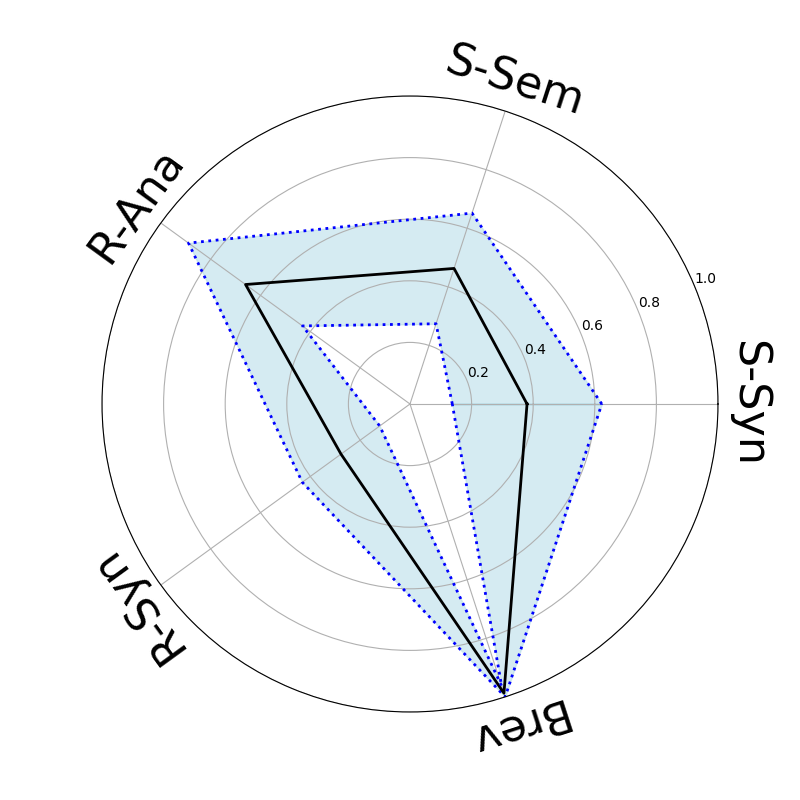}
        \caption{Llama 3 (8B)}
        \label{fig:compass-Llama-3--8B}
    \end{subfigure}
    \begin{subfigure}{0.23\textwidth}
        \includegraphics[width=\linewidth]{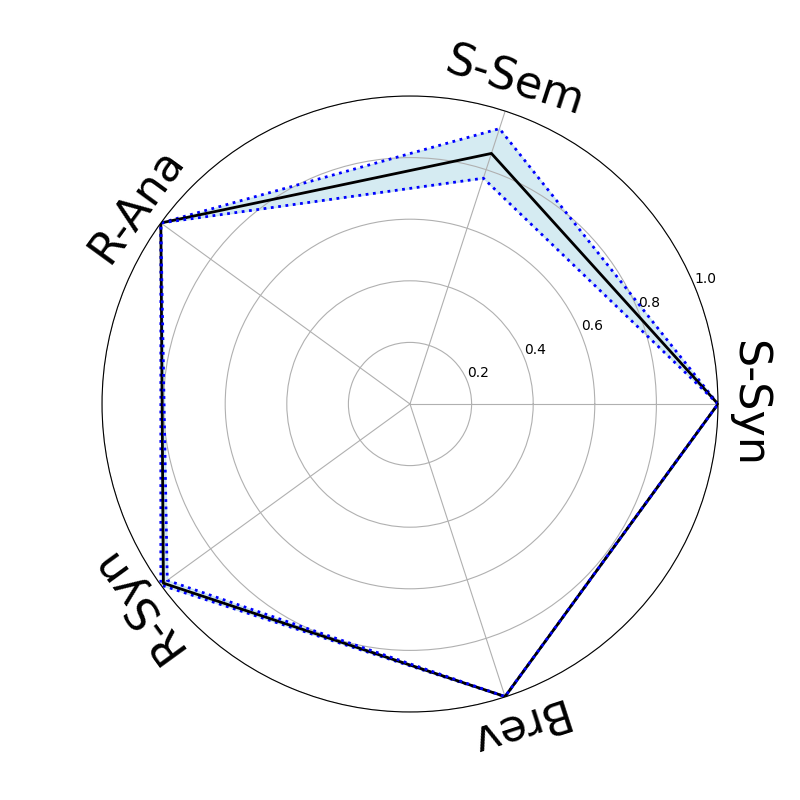}
        \caption{Qwen 2 72B}
        \label{fig:compass-Qwen-2--72B}
    \end{subfigure}
    \caption{Examples of capability compasses generated with the framework on the dataset.
     Five dimensions are configured in an exemplary way: Brevity(Brev), RDF Syntax (R-Syn), RDF Analytics(R-Ana), SPARQL Semantics(S-Sem) and SPARQL Syntax(S-Syn)}
    \label{fig:compasses}
\end{figure}

\subsubsection{Selection of Open LLMs}
We based our selection of state-of-the-art open LLMs on the Open LLM Leaderboard \cite{open-llm-leaderboard-v2} and used the average score over all included benchmarks as our reference value. The selection criteria were that the model is instruction-finetuned, as required by the task construction of the LLM-KG-bench framework, has less than 80B parameters because of a limited amount of available hardware resources and is a base model, i.e., not a fine-tuned version of another model. With the latter requirement, we wanted to stick to mature and popularly used LLMs that are not just optimized to achieve a slightly higher score on one or few benchmarks than a base model. 

The models fulfilling all our criteria and were among the TOP-4 models based on the average benchmark scores, disregarding models of the same family with lower scores, were Qwen2-72B-Instruct, Meta-Llama-3.1-70B-Instruct, solar-pro-preview-instruct and Phi-3.5-MoE-instruct. Here, we excluded solar-pro-preview-instruct from our selection since it only supports a context length of up to 4096k Tokens and not all prompts of tasks included in the run fitted within this limit. For the remaining three models, we also included all models of their larger model families that matched our requirements, i.e., we also tested all models of the Llama3 \cite{grattafiori2024llama3herdmodels}, Qwen2 \cite{hui2024qwen25codertechnicalreport,yang2024qwen2technicalreport}, and Phi3 \cite{abdin2024phi3technicalreporthighly} families fulfilling our requirements.

In addition, we wanted to test open LLMs that are fine-tuned or explicitly optimized on code since they could potentially better understand and produce structured data as required for the tasks included in the LLM-KG-Bench. Here, we consulted the EvalPlus Leaderboard \cite{liu2024your} and used the reported models' Mostly Basic Python Programming (MBPP) Benchmark score as our reference value to assess the code-producing quality of the models. Again, we excluded models that were only finetuned versions of a code-finetuned or -optimized base model, had more than 80B parameters or were not instruction-finetuned. Moreover, we only searched for models that are fine-tuned or explicitly optimized on code. Finally, we included the Top-3 models satisfying the criteria in our runs, namely Qwen2.5-Coder-32B-Instruct  \cite{hui2024qwen25codertechnicalreport}, DeepSeek-Coder-33B-Instruct \cite{guo2024deepseekcoderlargelanguagemodel} and OpenCoder-8B-Instruct \cite{huang2024opencoderopencookbooktoptier}.

\begin{table}[tb!]
    \caption{
    Result of two sided t-tests checking for a preference for Turtle vs JSON-LD serialization.
    Preferences are expected if the confidence interval is at least 95\%, bold font indicates 99\% or better.
    }
    \label{tab:TurtleVsJsonld}
    \scriptsize
    \centering
    \begin{tabular}{lcccccc}
    \toprule
        & \rot{RdfConnectionExplainStatic} & \rot{RdfFriendCount-1} & \rot{RdfFriendCount-2} & \rot{RdfSyntaxFixList} & \rot{Sparql2AnswerListOrga} & \rot{Text2AnswerListOrga} \\
    \midrule        
        GPT3.5 2024/01 & - & JSON & \textbf{JSON} & - & - & \textbf{TTL} \\
        GPT4o-mini 2024/07 & - & - & TTL & - & - & \textbf{TTL} \\
        GPT4o 2024/11 & - & - & - & JSON & - & - \\
        GPTo1-mini 2024/09 & - & - & - & - & - & - \\
        GPTo1-pre 2024/09 & - & - & - & - & - & - \\
        Gemini 1.5 Flash & - & \textbf{TTL} & - & - & - & - \\
        Gemini 1.5 Pro & - & - & - & - & - & - \\
        Gemini 2.0 Flash Exp & - & - & - & - & - & - \\
        Claude 3.5 Haiku & \textbf{JSON} & \textbf{TTL} & \textbf{TTL} & - & - & - \\
        Claude 3.5 Sonnet & - & - & - & - & - & - \\
        Qwen2-0.5B-Instruct & - & - & JSON & - & - & - \\
        Qwen2-1.5B-Instruct & \textbf{TTL} & \textbf{JSON} & \textbf{JSON} & \textbf{JSON} & - & - \\
        Qwen2-7B-Instruct & \textbf{JSON} & - & - & - & - & TTL \\
        Qwen2-57B-A14B-Instruct & - & - & - & \textbf{JSON} & TTL & \textbf{TTL} \\
        Qwen2-72B-Instruct & - & - & - & \textbf{JSON} & - & - \\
        Qwen2.5-0.5B-Instruct & - & - & - & - & - & - \\
        Qwen2.5-1.5B-Instruct & - & - & - & \textbf{JSON} & - & - \\
        Qwen2.5-3B-Instruct & \textbf{JSON} & - & - & \textbf{JSON} & - & - \\
        Qwen2.5-7B-Instruct & \textbf{JSON} & - & - & - & \textbf{TTL} & \textbf{TTL} \\
        Qwen2.5-14B-Instruct & \textbf{TTL} & - & - & - & - & TTL \\
        Qwen2.5-32B-instruct & - & \textbf{TTL} & \textbf{TTL} & \textbf{JSON} & \textbf{JSON} & - \\
        Qwen2.5-72B-Instruct & \textbf{JSON} & JSON & \textbf{JSON} & - & - & \textbf{TTL} \\
        Qwen2.5-Coder-32B-Instruct & - & \textbf{TTL} & \textbf{TTL} & - & - & - \\
        Meta-Llama-3-8B-Instruct & \textbf{TTL} & - & - & - & - & - \\
        Meta-Llama-3-70B-Instruct & - & \textbf{JSON} & \textbf{JSON} & \textbf{JSON} & - & - \\
        Meta-Llama-3.1-8B-Instruct & - & \textbf{JSON} & \textbf{JSON} & JSON & - & - \\
        Meta-Llama-3.1-70B-Instruct & - & - & - & \textbf{JSON} & \textbf{TTL} & - \\
        Meta-Llama-3.2-1B-Instruct & - & - & - & \textbf{JSON} & - & - \\
        Meta-Llama-3.2-3B-Instruct & \textbf{JSON} & \textbf{TTL} & \textbf{TTL} & \textbf{JSON} & - & - \\
        Meta-Llama-3.3-70B-Instruct & - & \textbf{JSON} & \textbf{JSON} & \textbf{JSON} & - & - \\
        Phi-3-mini-128k-instruct & - & - & \textbf{JSON} & \textbf{JSON} & - & - \\
        Phi-3-small-128k-instruct & \textbf{JSON} & - & - & TTL & - & - \\
        Phi-3-medium-128k-instruct & \textbf{TTL} & - & JSON & - & - & TTL \\
        Phi-3.5-mini-instruct & \textbf{TTL} & JSON & - & \textbf{JSON} & \textbf{TTL} & - \\
        OpenCoder-8B-Instruct & - & - & - & - & TTL & - \\
        Deepseek-Coder-33B-Instruct & - & \textbf{JSON} & \textbf{JSON} & \textbf{JSON} & - & - \\
    \bottomrule
    \end{tabular}
\end{table}

\subsection{Example Evaluations}
\label{sec:eval}
The generated data can be analysed in different ways.
The raw data contains all LLM interactions, extensive logs and the evaluation results in json, yaml and txt format including interaction and task details.
The \verb|plotResults| command helps to generate several boxplots (which we have omitted here due to limited space) as well as generate csv and excel files showing the results in big tables.

In \LKGBf Version 3, we added the capability to aggregate results for each model evaluated and create capability compass plots. We used an exemplary configuration to create the ones shown in \cref{fig:compasses}.
These plots can be used to give a summary of a model or create model cards.

Finally, we used the added task versions with different graph serialization formats to check if some models prefer JSON-LD over turtle serialization.
Table \ref{tab:TurtleVsJsonld} shows the result of two sided t-tests for all combinations of related tasks and all models.
A table cell contains "TTL" or "JSON" if the statistics indicate better scores for one format.
Bold font is used if the confidence is at least 99\%.

\section{Conclusion and Future Work}
\label{sec:conclusions}
In this work, we presented the updated \LKGBf in Version 3 which features new tasks, features and capabilities.
Our framework complements the landscape of LLM benchmarks as many major frameworks do not specifically take semantic aspects into account. In the evaluation, we used our framework to assess the semantic capabilities of more than 30 LLMs. The results were collected in a large dataset and the data is published to facilitate further analysis and comparison.
One line of future work is directed at the engineering of new tasks and test cases for the automated evaluation of KGE-related capabilities. The other at better enabling further in-depth analysis of the evaluation results, thereby integrating existing scoring methods or devising new ones.
We released the \LKGBf as a means to study LLMs and to enable collaboration with researchers on this major task. We are looking forward to contributions. 

\subsection*{Online Resources}
\label{sec:OnlineResources}
The framework source code and generated result datasets are publicly available:
\
\paragraph{LLM-KG-Bench Framework Code:} ~\\
\url{https://github.com/aksw/LLM-KG-Bench} 
and \href{https://doi.org/10.5281/zenodo.15100803}{DOI:10.5281/zenodo.15100803} 

\paragraph{Generated Dataset:} 
\url{https://github.com/AKSW/LLM-KG-Bench-Results/tree/main/2025-ESWC_LLM-KG-Bench-3-Results}
and \href{https://doi.org/10.5281/zenodo.15100645}{DOI:10.5281/zenodo.15100645} 
\paragraph{All Results Published So Far:} ~\\
\url{https://github.com/AKSW/LLM-KG-Bench-Results}

\begin{credits}
\subsubsection{\ackname}
We thank DFKI for running the experiments involving open LLMs on their GPU cluster.
This work was partially supported by grants from the German Federal Ministry of Education and Research (BMBF) to the projects ScaleTrust (16DTM312D) and KupferDigital2 (13XP5230L) as well as from the German Federal Ministry for Economic Affairs and Climate Action (BMWK) to the KISS project (01MK22001A) and CoyPu project (01MK21007A) and from the German Federal Ministry of Transport and Digital Infrastructure (BMDV) to the project ADA (19F2190B).

\subsubsection{\discintname}
The authors used free Google Cloud credits for the execution of Gemini models, however due to the setup and technical nature of the evaluation this has no effect on the results.
\end{credits}
%
%
%
\bibliographystyle{splncs04}
\bibliography{ref}

\clearpage
\pagestyle{plain}
\cfoot*{} 

\section*{Metadata for this article}

\begin{description}
    \item[Title:] LLM-KG-Bench 3.0: A Compass for Semantic Technology Capabilities in the Ocean of LLMs
    \item[Authors:] Lars-Peter Meyer, Johannes Frey*, Desiree Heim*, Felix Brei*, Claus~Stadler, Kurt Junghanns and Michael Martin\\
      ~* authors contributed equally
    \item[conference:] \href{https://2025.eswc-conferences.org/}{ESWC 2025}, 1.-5.~6.~2025 in Portoroz, Slovenia
    \item[submitted for review:] 20.~12.~2024
    \item[peer review status:] accepted by peer review (25.~2.~2025)
    \item[submitted camera ready:] 28.~3.~2025
    \item[publication date:] about 2025
    \item[LLM-KG-Bench Framework Code:] ~
      \url{https://github.com/aksw/LLM-KG-Bench} and \href{https://doi.org/10.5281/zenodo.15100803}{DOI:10.5281/zenodo.15100803}
    \item[Generated Dataset:] 
      \url{https://github.com/AKSW/LLM-KG-Bench-Results/tree/main/2025-ESWC_LLM-KG-Bench-3-Results} and \href{https://doi.org/10.5281/zenodo.15100645}{DOI:10.5281/zenodo.15100645} 
    \item[Bibtex entry:] 
\end{description}

\begin{scriptsize}
  \begin{verbatim}
@inproceedings{Meyer2025LlmKgBench3,
  author = {L.-P. Meyer and J. Frey and D. Heim and F. Brei and C. Stadler and K. Junghanns and M. Martin},
  title = {{LLM-KG-Bench} 3.0: A Compass for Semantic Technology Capabilities in the Ocean of {LLMs}},
  year = {2025},
  booktitle = {The Semantic Web, 22nd International Conference, {ESWC} 2025, Proceedings}
}    
  \end{verbatim}
\end{scriptsize}

\end{document}